\pdfoutput=1

\documentclass[11pt]{article}

\usepackage[]{acl}

\usepackage{times}
\usepackage{latexsym}

\usepackage[T1]{fontenc}

\usepackage[utf8]{inputenc}

\usepackage{microtype}

\usepackage{booktabs}
\usepackage{xcolor}
\usepackage{multirow}
\usepackage{setspace}
\usepackage{pdflscape}
\usepackage{enumitem}
\usepackage{dirtytalk}
\usepackage{subfigure}
\usepackage{wrapfig}
\usepackage{danudefs} 

\usepackage{mathtools}
\usepackage{times}
\usepackage{latexsym}
\usepackage{amsfonts,amsmath,amssymb,bm}
\usepackage{times}
\usepackage{multirow}
\usepackage{graphicx}
\usepackage{xurl} 
\usepackage[english]{babel}
\usepackage{hyperref}
\addto\captionsenglish{%
}
\addto\extrasenglish{%
}

\title{\emph{Debiasing isn't enough!} -- On the Effectiveness of Debiasing MLMs and their Social Biases in Downstream Tasks}

\author{Masahiro Kaneko$^{1}$ \quad
        Danushka Bollegala$^{2,3}$\Thanks{ Danushka Bollegala holds concurrent appointments as a Professor at University of Liverpool and as an Amazon Scholar. This paper describes work performed at the University of Liverpool and is not associated with Amazon.} \quad
        Naoaki Okazaki$^{1}$ \\
        $^1$Tokyo Institute of Technology \quad
        $^2$University of Liverpool \quad
        $^3$Amazon \\
        {\tt masahiro.kaneko@nlp.c.titech.ac.jp} \\
        {\tt danushka@liverpool.ac.uk} \quad
        {\tt okazaki@c.titech.ac.jp}
}

\begin{document}
\maketitle
\begin{abstract}


We study the relationship between task-agnostic intrinsic  and task-specific extrinsic social bias evaluation measures for Masked Language Models (MLMs), and 
find that there exists only a weak correlation between these two types of evaluation measures.
Moreover, we find that MLMs debiased using different methods still re-learn social biases during fine-tuning on downstream tasks.
We identify the social biases in both training instances as well as their assigned labels as reasons for the discrepancy between intrinsic and extrinsic bias evaluation measurements.
Overall, our findings highlight the limitations of existing MLM bias evaluation measures and raise concerns on the deployment of MLMs in downstream applications using those measures.
\end{abstract}

\section{Introduction}
\label{sec:intro}

Text representations produced by MLMs have revolutionised NLP by improving the performance of numerous downstream applications~\cite{Elmo,BERT,RoBERTa,ALBERT,XLNet}.
Unfortunately, large-scale pretrained MLMs demonstrate worrying levels of social biases when applied to downstream tasks~\cite{may-etal-2019-measuring,nadeem-etal-2021-stereoset,Kaneko:EACL:2021b,Kaneko:MLM:2022}.
Given that real-world Natural Language Processing (NLP) systems such as machine translation systems, dialogue systems, etc. are used by millions of users world-wide~\cite{Hovy:ACL:2016}, it remains an important responsibility to accurately evaluate the social biases in MLMs prior to deployment.

Two types of social bias evaluation measures have been proposed for MLMs in prior work~\cite{goldfarb-tarrant-etal-2021-intrinsic,Cao:2022}: (a) task-agnostic \emph{intrinsic} evaluation measures~\cite{crows-pairs,nadeem-etal-2021-stereoset,Kaneko:AAAI2022} that use the likelihood scores assigned by an MLM under evaluation for sentences representing various social biases,
and (b) task-specific \emph{extrinsic} evaluation measures that use the data from downstream NLP tasks such as predicting the occupation of a person from their biographies~\cite{bias-in-bios,bartl-etal-2020-unmasking}.
In contrast to intrinsic measures that directly probe into the biases in MLMs, extrinsic measures holistically evaluate an NLP system that uses an MLM.

Much prior work has decoupled the intrinsic and extrinsic bias evaluations for simplicity, and have assumed that an MLM unbiased according to an intrinsic measure will remain to be so under extrinsic evaluations too.
However, this unverified hypothesis begs the following question: \emph{Can we reliably predict the social biases of an MLM when it is applied to a particular downstream task using only intrinsic evaluation measures?}
To answer this question, we conduct a comprehensive study using 7 pretrained MLMs and their debiased versions using three different debiasing methods.
However, only weak correlations are found between intrinsic and extrinsic bias evaluation measures.
This is a worrying proposition because intrinsic measures are often used to decide whether an MLM should be deployed in a downstream application.

We further investigate on why MLMs learn social biases during downstream task fine-tuning and identify two sources: \emph{instance-related} biases and \emph{label-related} biases.
The training instances for downstream tasks contain unfair associations that are learnt by the masked language modelling objective, which we name instance-related biases.
We find that debiased MLMs gradually learn such biases during fine-tuning, leading to disagreements between intrinsic and extrinsic evaluation measures.
Moreover, the labels assigned to downstream training instances can also represent biased ratings, which are also learnt by the MLMs when their parameters are updated in an end-to-end manner when predicting task labels.
Overall, we find weak correlations between intrinsic and extrinsic measures even after the models are fine-tuned on downstream task data.
Based on our findings, we recommend that intrinsic bias evaluation measure alone must not be used to determine whether an MLM should be deployed in a downstream NLP application.


\section{Related Work}
\label{sec:related}


Social biases have been reported in models trained for numerous downstream tasks.
\newcite{kiritchenko-mohammad-2018-examining} examined 219 automatic sentiment analysis systems that participated in the SemEval-2018 Task 1~\cite{mohammad-etal-2018-semeval} and found that many systems show statistically significant gender or racial biases.
\newcite{Diaz:2019} investigated the age-related biases in sentiment classification and found that across 15 sentiment analysis tools, sentences containing adjectives describing the youth were 66\% more likely to be scored positively than those describing the elderly.
The problem is also observed in MLMs trained for many languages~\cite{Kaneko:MLM:2022}.

Numerous bias mitigation approaches have been proposed for MLMs such as fine-tuning~\cite{Kaneko:EACL:2021b,lauscher-etal-2021-sustainable-modular}, counterfactual data augmentation~\cite{hall-maudslay-etal-2019-name,Zmigrod:2019} and parameter dropout~\cite{webster2020measuring}.
\newcite{liang-etal-2020-towards} proposed a method to debias the sentence embeddings created from BERT~\cite{BERT} and ELMo~\cite{Elmo} inspired by hard debiasing ~\cite{Tolga:NIPS:2016}, a method originally proposed for static word embeddings.
However, not all biases are adequately mitigated by the current proposals~\cite{Gonen:2019}.
Our focus in this paper is the evaluation and not debiasing methods.

Intrinsic bias evaluation measures~\cite{crows-pairs,nadeem-etal-2021-stereoset,Kaneko:AAAI2022} evaluate the social biases in a given MLM standalone, independently of any downstream applications.
Pseudo log-likelihood scores assigned to stereotypical vs. antistereotypical examples (either manually written or automatically generated using templates) have been used to evaluate the social biases in an MLM as further detailed in  \autoref{sec:bias-measures}.
Considering that MLMs are used to represent input texts in various downstream tasks, several prior work have argued that their social biases must be evaluated with respect to those tasks~\cite{bias-in-bios,bartl-etal-2020-unmasking,webster2020measuring}.
However, decoupling tasks and MLMs makes the bias evaluation simpler and task-independent, which is attractive because MLMs are designed as generic text representations for a wide-range of downstream tasks.
We further discuss extrinsic bias evaluation measures in \autoref{sec:bias-measures}.
\newcite{goldfarb-tarrant-etal-2021-intrinsic} studied social biases in static word embeddings using the Word Embedding Association Test (WEAT) as an intrinsic measure and
coreference detection and hate speech detection as extrinsic evaluation tasks.
They found the correlations between intrinsic and extrinsic measures to be weak or negative.
Although we share the motivation with \newcite{goldfarb-tarrant-etal-2021-intrinsic}, our focus is contextualised embeddings produced by MLMs, which have outperformed static word embeddings in numerous tasks, hence more widely used in real-world NLP applications than static word embeddings.

\newcite{Cao:2022} studied the correlation between intrinsic bias scores (i.e. Contextualised Embedding Association Test~\cite[][CEAT]{CEAT} and Increased Log Probability Score~\cite[][ILPS]{kurita-etal-2019-measuring}) and extrinsic bias scores for three tasks: toxicity~\cite{Jigsaw}, hate-speech~\cite{Mathew2020-nt} and sentiment~\cite{BOLD} detection.
Similar to \newcite{goldfarb-tarrant-etal-2021-intrinsic} they also found weak correlations.
As reasons for this lack of correlation they highlight misalignment between the metrics such as the notion of bias, protected groups, and noise in the evaluation datasets.
Our contributions are complementary to theirs because we consider different intrinsic and extrinsic measures, not covered by them.
Moreover, our extrinsic tasks are directly related to social biases.
We observe weak correlations between intrinsic and extrinsic evaluation measures, supporting their claims.

\section{Bias Evaluation Measures}
\label{sec:bias-measures}

Several measures have already been proposed in prior work for evaluating the social biases encoded in MLMs.
These measures can be broadly categorised into two groups depending on whether they evaluate social biases in MLMs with respect to a particular downstream task or not. 
We refer to measures that evaluate social biases in MLMs on their own right, independently of any downstream tasks as \emph{intrinsic MLM bias scores} and describe three such measures. 
In contrast, we refer to the measures that evaluate social biases in MLMs when they are applied to solve a specific downstream task such as Natural Language Inference (NLI), Semantic Textual Similarity (STS) or predicting occupations from biographies as \emph{extrinsic MLM bias scores} and describe those measures.


To describe the intrinsic measures, let us consider a test sentence $S = w_1, w_2, \ldots, w_{|S|}$, containing length $|S|$ sequence of tokens $w_i$, where part of $S$ is modified to create a stereotypical (or lack of thereof) example for a particular social bias. 
For example, let's consider the sentence-pair ``\emph{\textbf{John} completed \textbf{his} PhD in machine learning}''  vs. ``\emph{\textbf{Mary} completed \textbf{her} PhD in machine learning}''.
Here, \{\emph{John}, \emph{his}\} are the modified tokens for the first sentence,  whereas for the second sentence they are \{\emph{Mary}, \emph{her}\}.
The unmodified tokens between  the two sentences are \{\emph{completed}, \emph{PhD}, \emph{in}, \emph{machine}, \emph{learning}\}.

Let us denote the list of modified tokens in a given sentence $S$ by $M$ and the list of unmodified tokens by $U$, so that $S = M \cup U$ is the list of all tokens in $S$.
\footnote{Note that to account for repeated occurrences of a word in a sentence, we examine lists rather than sets.}
Given an MLM with pre-trained parameters $\theta$ to evaluate for social biases, let us denote the probability $P_{\mathrm{MLM}}(w_i | S_{\setminus w_{i}}; \theta)$ that the MLM assigns to a token $w_i$ conditioned on the remainder of the tokens, $S_{\setminus w_{i}}$.
\citet{salazar-etal-2020-masked} demonstrated that $\mathrm{PLL}(S)$, the pseudo-log-likelihood (PLL) score of sentence $S$ given by \eqref{eq:PLL}, can be used to evaluate the preference expressed by an MLM for $S$, similarly to using log-probabilities for evaluating the naturalness of sentences using conventional language models.
\begin{align}
    \label{eq:PLL}
    \mathrm{PLL}(S) \coloneqq \sum_{i=1}^{|S|} \log P_{\mathrm{MLM}}(w_i | S_{\setminus w_{i}}; \theta)
\end{align}
PLL scores for MLMs can be computed directly and are more uniform across sentence lengths (no left-to-right bias), allowing us to recognize natural sentences in a language~\cite{wang-cho-2019-bert}.
As we will see later, PLL can be used to define bias evaluation scores for MLMs in a variety of ways.

\paragraph{Intrinsic: StereoSet Score}
The probability of generating the modified tokens given the unmodified tokens in $S$ was expressed as $P(M | U; \theta)$ by \citet{Nadeem:2020}.
We refer to this as StereoSet Score (\textbf{SSS}) and it is given by \eqref{eq:SSS}.
\begin{align}
    \label{eq:SSS}
    \mathrm{SSS}(S) \coloneqq  \frac{1}{|M|} \sum_{w \in M} \log P_{\mathrm{MLM}}(w | U; \theta)
\end{align}
$|M|$ denotes the length of $M$.
SSS presents a challenge because, when comparing $P(M | U; \theta)$ for modified words like \emph{John}, we might see high probabilities simply because these words occur frequently in the data used to train the MLM and not because the MLM has picked up a social bias.

\paragraph{Intrinsic: CrowS-Pairs Score} The CrowS-Pairs Score (\textbf{CPS}) is a scoring formula provided by \eqref{eq:CPS} that was defined as $P(U|M; \theta)$ by \citet{crows-pairs} to address the frequency-bias in SSS.
\begin{align}
    \label{eq:CPS}
     \mathrm{CPS}(S) \coloneqq \sum_{w \in U} \log P_{\mathrm{MLM}}(w | U_{\setminus w}, M; \theta)
\end{align}
Normalization is not performed here because the length of unmodified tokens is the same.
However, by masking and predicting one token $w$ at a time from $U$, we are effectively changing the context $(U_{\setminus w}, M)$ used as input by the MLM.
This has two disadvantages.
First, by removing $w$ from the sentence, the MLM loses information that it can use to predict $w$.
As a result, the prediction accuracy of $w$ may decrease, rendering bias evaluations untrustworthy.
Second, the remaining tokens $\{U_{\setminus w}, M\}$ may still be biased even if we remove one token $w$ at a time from $U$.
Moreover, the context on which we condition the probabilities continuously varies across predictions.

\paragraph{Intrinsic: All Unmasked Likelihood with Attention weights} \citet{Kaneko:AAAI2022} proposed All Unmasked Likelihood with Attention weights (\textbf{AULA}) to overcome the above-mentioned disfluencies in previously proposed MLM bias evaluation measures.
To begin, rather than masking out tokens from $S$, AULA provides the entire sentence to the MLM.
Second, all tokens in $S$ that appear in between the start and end of sentences are predicted by AULA.
Furthermore, AULA computes the likelihood using attention weights to evaluate social biases based on the relative importance of words in a sentence, as given by \eqref{eq:AULA}.
\begin{align}
    \label{eq:AULA}
    \mathrm{AULA}(S) \coloneqq \frac{1}{|S|} \sum_{i=1}^{|S|} \alpha_i \log P_{\mathrm{MLM}}(w_i | S; \theta)
\end{align}
Here, $\alpha_i$ represents the average of all multi-head attentions associated with $w_i$.


Given a score function $f \in \{\mathrm{SSS}, \mathrm{CPS}, \mathrm{AULA}\}$, we use the percentage of stereotypical ($S^{st}$) test sentences preferred by the MLM over anti-stereotypical ($S^{at}$) ones to define the corresponding bias evaluation measure (\textbf{bias score}) as follows:
\begin{align}
  \label{eq:score}
  \left( \frac{100}{N}\sum_{(S^{\rm st}, S^{\rm at})} \mathbb{I}(f(S^{\rm st}) > f(S^{\rm at}))\right) - 50
\end{align}
Here, the indicator function $\mathbb{I}$ returns $1$ if its argument is True and $0$ otherwise, and $N$ is the total number of test instances.
Values close to 0 indicate that the MLM under consideration is neither stereotypically nor anti-stereotypically biased; thus, it can be considered unbiased.
Values less than 0 indicate a bias toward the anti-stereotypical group, while values greater than 0 indicate a bias toward the stereotypical group.

\paragraph{Extrinsic: BiasBios} To evaluate the gender bias in occupation classification, \newcite{bias-in-bios} proposed a dataset where they collected biographies written in English from Common Crawl.
For this purpose, they first filter lines that begin with a name-like-pattern (i.e. a sequence of two capitalised words) followed by the string ``is a(n) (xxx) \emph{title}'', where \emph{title} is an occupation from the BLS Standard Occupation Classification system.\footnote{\url{https://www.bls.gov/soc}}
They used twenty-eight frequent occupations in Common Crawl to collect 397,340 biographies.
The task is to predict the people's occupations, taken from the first sentence of their biographies, given the remainder of their biographies.
For example, given the hypothetical biography of a nurse:
\emph{She graduated from Lehigh University, with honours in 1998. Nancy has years of experience in weight loss surgery, patient support, education, and diabetes.}, the goal is to predict the occupation.

An MLM under evaluation is used to create an embedding for a biography excluding the title sentence, and a softmax classifier is trained using those biography embeddings to predict the occupation of the person described in each biography.
If the MLM is unbiased with respect to gender, we would expect it to have approximately similar true positive rates (TPRs) averaged over all twenty-eight occupations for both male and female gendered persons.
The TPR gap between the TPRs for classifying occupations in female and male biographies is used as an extrinsic MLM bias score.
An unbiased MLM is expected to predict occupations with similar TPRs for both genders.

\paragraph{Extrinsic: STS-bias}
Semantic Textual Similarity (STS) is a task which evaluates the degree of semantic similarity between two sentences.
\citet{webster2020measuring} proposed STS-bias to evaluate gender bias by adapting STS formulation.
Specifically, to create the STS-bias dataset, they collect the 276 sentences from the STS-B test set, which start with \textit{A man} or \textit{A woman}, and discard the sentences with multiple gendered words.\footnote{Gender word list: \url{https://github.com/uclanlp/gn_glove/tree/master/wordlist}}
For each template, they formed two sentence pairs per profession from \citet{rudinger-etal-2018-gender}, one representing the male gender and the other for the female gender.
For example, given a pair ``\textit{A man is walking}'' and ``\textit{A nurse is walking}'' from STS-B dataset, they create a new pair ``\textit{A woman is walking}'' and  ``\textit{A nurse is walking}''.
If not relying on gendered correlations, a model should assign equal estimates of similarity to the two pairs.
Therefore, following \citet{rudinger-etal-2017-social}, as a score we use the difference between similarities of such two pairs to measure how similar model predictions actually are regarding gender.
They compute the Pearson correlation between the scores per profession and the representation in that profession listed in the US Bureau of Labor Statistics~\cite{Caliskan2017SemanticsDA}.
STS-bias does not consider the directionality of the gender bias. 
Therefore, we report absolute values of all bias scores. 



\paragraph{Extrinsic: NLI-bias}
\newcite{Dev:2019} used natural language inference (NLI) for evaluating gender bias.
This dataset contains sentence triples where a premise must be neutral in entailment w.r.t. two hypotheses.
For example, ``\textit{The accountant ate a bagel}'' should neither entail nor contradict the sentences ``\textit{The man ate a bagel}'' and ``\textit{The woman ate a bagel}''.
Given a set containing $\cD$ test instances, let the entailment predictor's probabilities for the $d$-th instance for entail, neutral and contradiction labels be respectively $e_{d}$, $n_{d}$ and $c_{d}$.
Then, they proposed the measure Fraction Neutral (FN): ${\rm FN} = \frac{1}{|\cD|} \sum_{d=1}^{\cD} \textbf{1} [{\rm neutral} = {\rm max}(e_{d}, n_{d}, c_{d})]$.
For an ideal (bias-free) embedding, FN measure would be 1.
They compute the FN for each female and male sentences and use the difference between them as the bias score in the NLI-bias.

\section{Experiments and Findings}
\label{sec:exp}

\subsection{Datasets}

We used Crows-Pairs dataset~\cite{crows-pairs} as CPS and AULA evaluation data, and used StereoSet dataset~\cite{nadeem-etal-2021-stereoset} as SSS evaluation data.
In BiasBios\footnote{\url{https://github.com/microsoft/biosbias}}, we use the average performance gap in the male and female groups aggregated across all occupations to measure the bias, following~\citet{zhao2020gender}.
We crawled BiasBios data using publicly available code from the original authors, and used 80\% of the collected data as a training data, 10\% as development data, and 10\% as evaluation data.
Because the STS-bias dataset is not publicly available, we created the dataset following to the procedure proposed by~\newcite{webster2020measuring}.
We used STS-B as training and development data~\cite{cer-etal-2017-semeval}\footnote{\url{github.com/facebookresearch/SentEval}} for STS-bias, following \citet{webster2020measuring}.
We used the Multi-Genre Natural Language Inference (MNLI) as training data and development data~\cite{williams-etal-2018-broad}\footnote{\url{cims.nyu.edu/~sbowman/multinli/}} for NLI-bias~\footnote{\url{https://github.com/sunipa/On-Measuring-and-Mitigating-Biased-Inferences-of-Word-Embeddings}}, following ~\citet{Kaneko:EACL:2021b}.


\subsection{Debiasing methods}
\label{sec:debiasing}
We used three previously proposed methods that are widely used for debiasing MLMs such as All-layer Token-level debiasing~\cite[\textbf{AT};][]{Kaneko:EACL:2021b}, Counterfactual Data Augmentation debiasing~\cite[\textbf{CDA};][]{webster2020measuring}, and dropout debiasing~\cite[\textbf{DO};][]{webster2020measuring}.
These methods represent conceptually different approaches for debiasing and can be applied to any pretrained MLM model/algorithm.

\paragraph{AT debiasing:}
\citet{Kaneko:EACL:2021b} proposed a method for debiasing MLMs via fine-tuning by simultaneously 
(a) preserving the semantic information in the pretrained  MLM, and
(b) removing discriminative gender-related biases via an orthogonal projection in the hidden layers by operating at token- or sentence-levels, enabling debiasing on different layers and levels in the pre-trained contextualised embedding model.
It is independent of model architectures and pre-training methods, thus can be adapted to a wide range of MLMs.
They showed that applying token-level debiasing for all tokens and across all layers of an MLM produces the best performance.
We perform an additional phase of fine-tuning using AT debiasing.

\paragraph{CDA debiasing:}
CDA is a debiasing strategy to mitigate social bias based on data~\cite{webster2020measuring}.
CDA involves re-balancing a corpus by swapping bias attribute words in a dataset.
For example, the sentence ``\textit{the doctor treated his patient}'' could be changed to ``\textit{the doctor treated her patient}''.
The re-balanced corpus is then often used for further training to debias a model.
We experiment with debiasing pre-trained MLMs by performing an additional phase of fine-tuning on counterfactually augmented sentences.

\paragraph{DO debiasing}
\citet{webster2020measuring} proposed dropout regularization for a bias mitigation.
They increased the dropout parameters for MLM’s attention weights and hidden activations and further pre-trained them.
They showed that increased dropout regularization reduces gender bias within these MLMs.
They hypothesize that dropout’s interruption of the attention mechanisms within MLMs prevents MLMs from learning undesirable associations between words. 
Similar to AT and CDA, we perform an additional phase of fine-tuning using increased dropout regularization.

We used news-commentary-v15\footnote{\url{https://data.statmt.org/news-commentary/v15/}} corpus as additional training data for debiasing.
We used publicly available code\footnote{\url{https://github.com/kanekomasahiro/context-debias}} and default hyperparameters for AT debiasing.
We used~\citet{zhao-etal-2018-learning}'s word list to replace a feminine word to the corresponding masculine word and vice versa in sentences in the training data for CDA debiasing.
We set the dropout ratio to 0.15 for attention probabilities and  0.2 for all fully connected layers in DO debiasing.
Hyperparametors of CDA and DO debiasing are set to their default values in the Huggingface Transformers library \cite{wolf-etal-2020-transformers}.

\subsection{Models and Hyperparameters}
In our experiments, we used the following seven MLMs: bert-base-uncased (\textbf{bert-bu})\footnote{\url{huggingface.co/bert-base-uncased}},
bert-base-cased (\textbf{bert-bc}),\footnote{\url{huggingface.co/bert-base-cased}}
bert-large-uncased (\textbf{bert-lu}),\footnote{\url{huggingface.co/bert-large-uncased}}
bert-large-cased (\textbf{bert-lc}),\footnote{\url{huggingface.co/bert-large-cased}}
roberta-base (\textbf{roberta-b}),\footnote{\url{huggingface.co/roberta-base}}
roberta-large (\textbf{roberta-l}),\footnote{\url{huggingface.co/roberta-large}}
albert-base (\textbf{albert-b}).\footnote{\url{huggingface.co/albert-base-v2}}

To fine-tune an MLM for a downstream task, we first create an embedding for a training instance using the given MLM.
Specifically, we use the [CLS] token embedding produced by the MLM to represent an instance.
For STS and NLI tasks where an instance consists of a pair of sentences, we concatenate the two sentences in the pair and use it as the input to the MLM.
Finally, a classification head (i.e. a softmax layer) is used to predict the label for a training instance, and cross entropy error is backpropagated to update the MLM parameters.

For downstream tasks, the best performance checkpoint on development data is selected from $\{16, 32, 64\}$ for batch size, \{1e-5, 3e-5, 5e-5\} for learning rate, $\{1, 3, 5\}$ for epoch number with greedy search.
Maximum sentence length is set to 128 tokens and we used four Tesla V100 GPUs in our experiments.
All other hyperparameters, are set to their default values in the huggingface library.

\subsection{Debiasing vs.  Bias in Downstream Tasks}
\label{sec:downstream}

\begin{figure*}[t!]
    \centering
    \subfigure[SSS]{\includegraphics[clip, width=0.32\textwidth]{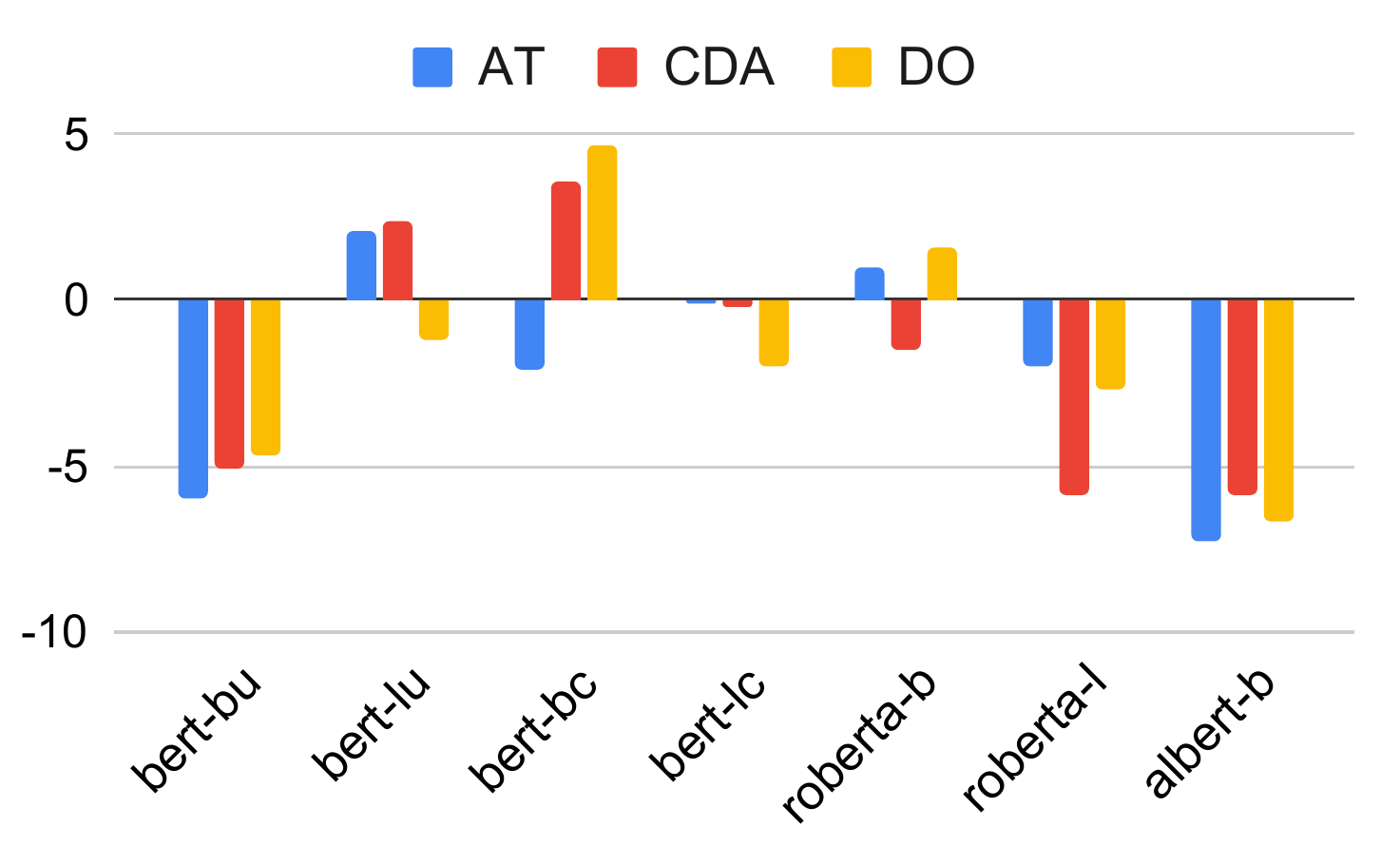}}
    \subfigure[CPS]{\includegraphics[clip, width=0.32\textwidth]{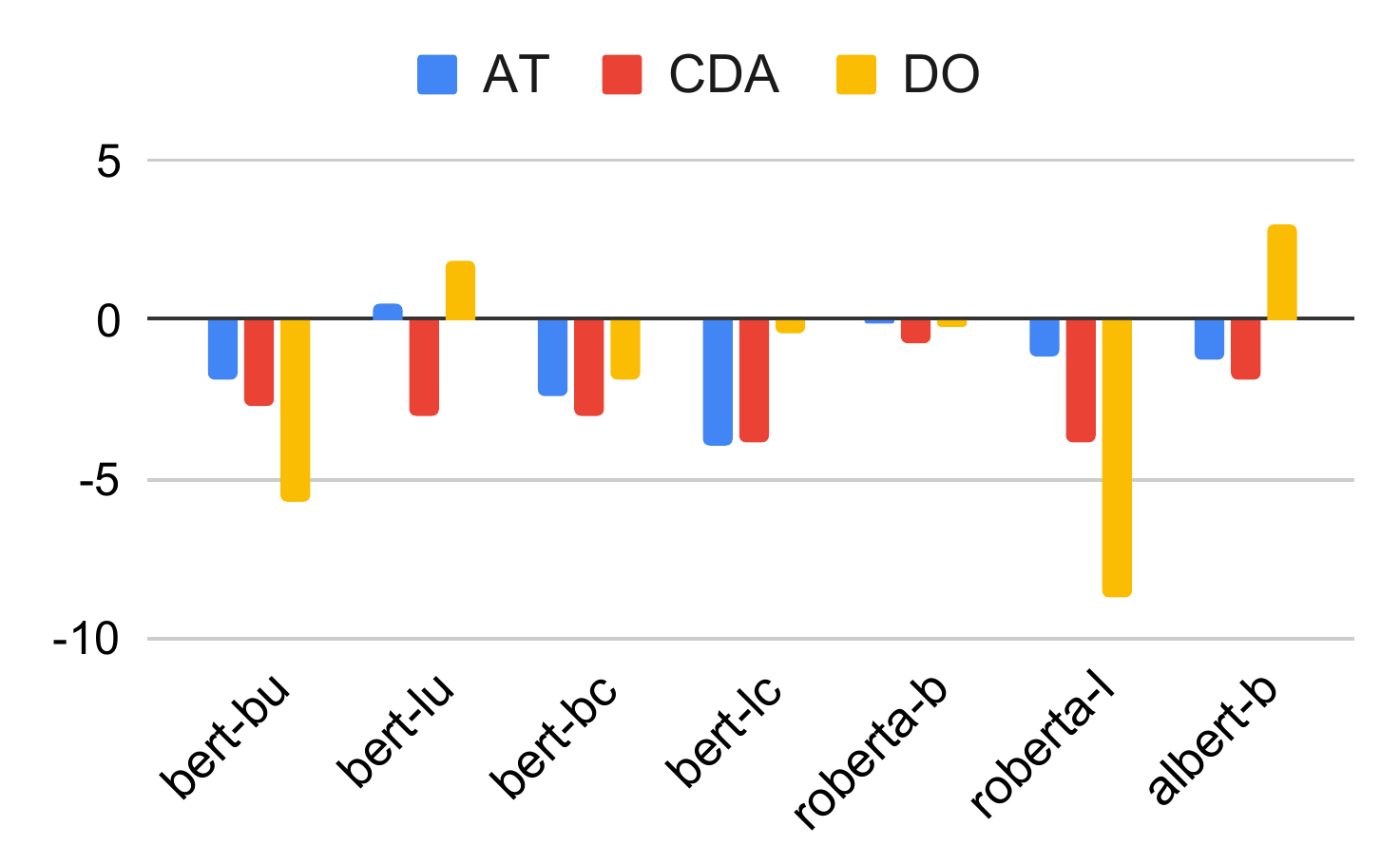}}
    \subfigure[AULA]{\includegraphics[clip, width=0.32\textwidth]{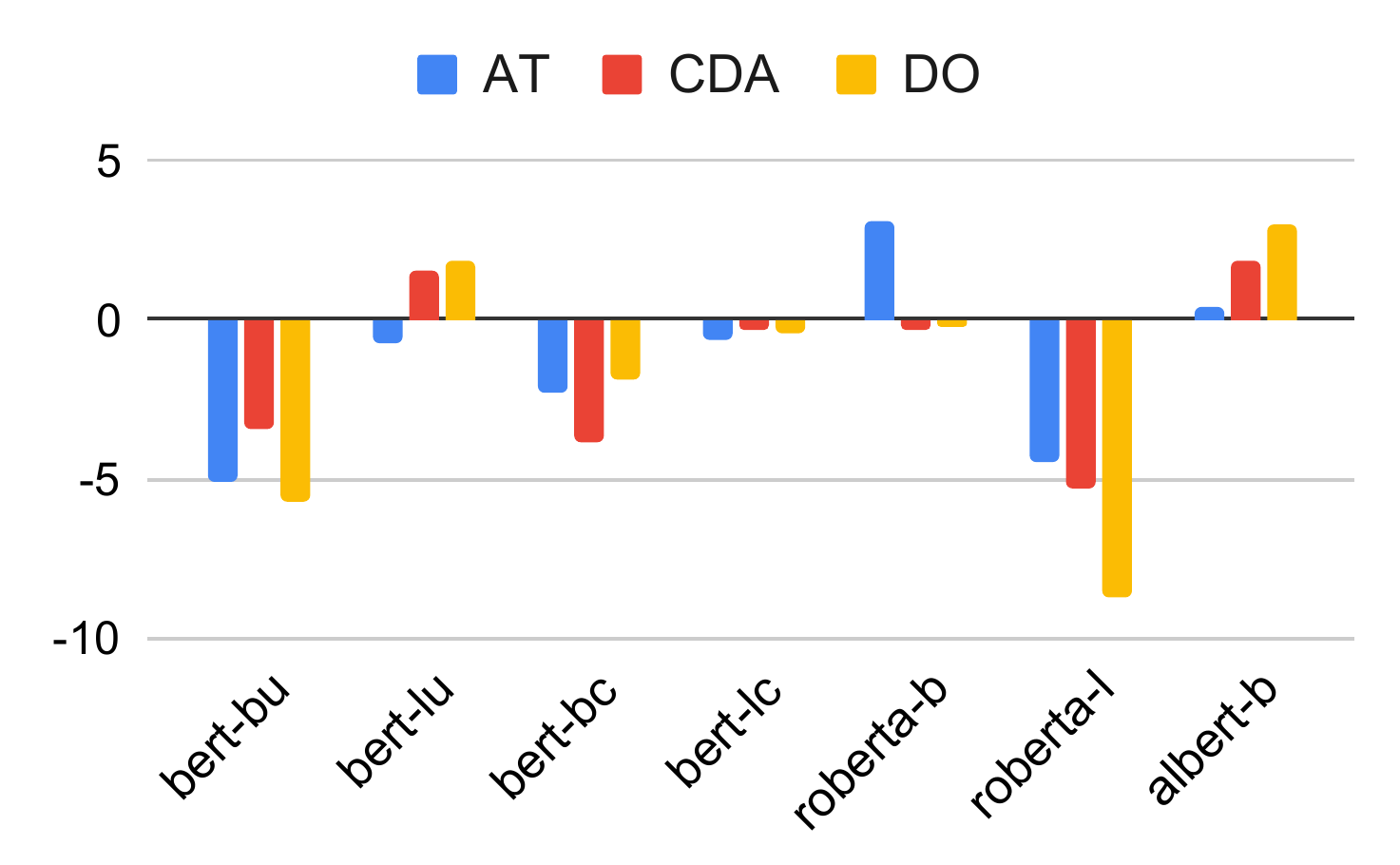}}
    
    \subfigure[BiasBios]{\includegraphics[clip, width=0.32\textwidth]{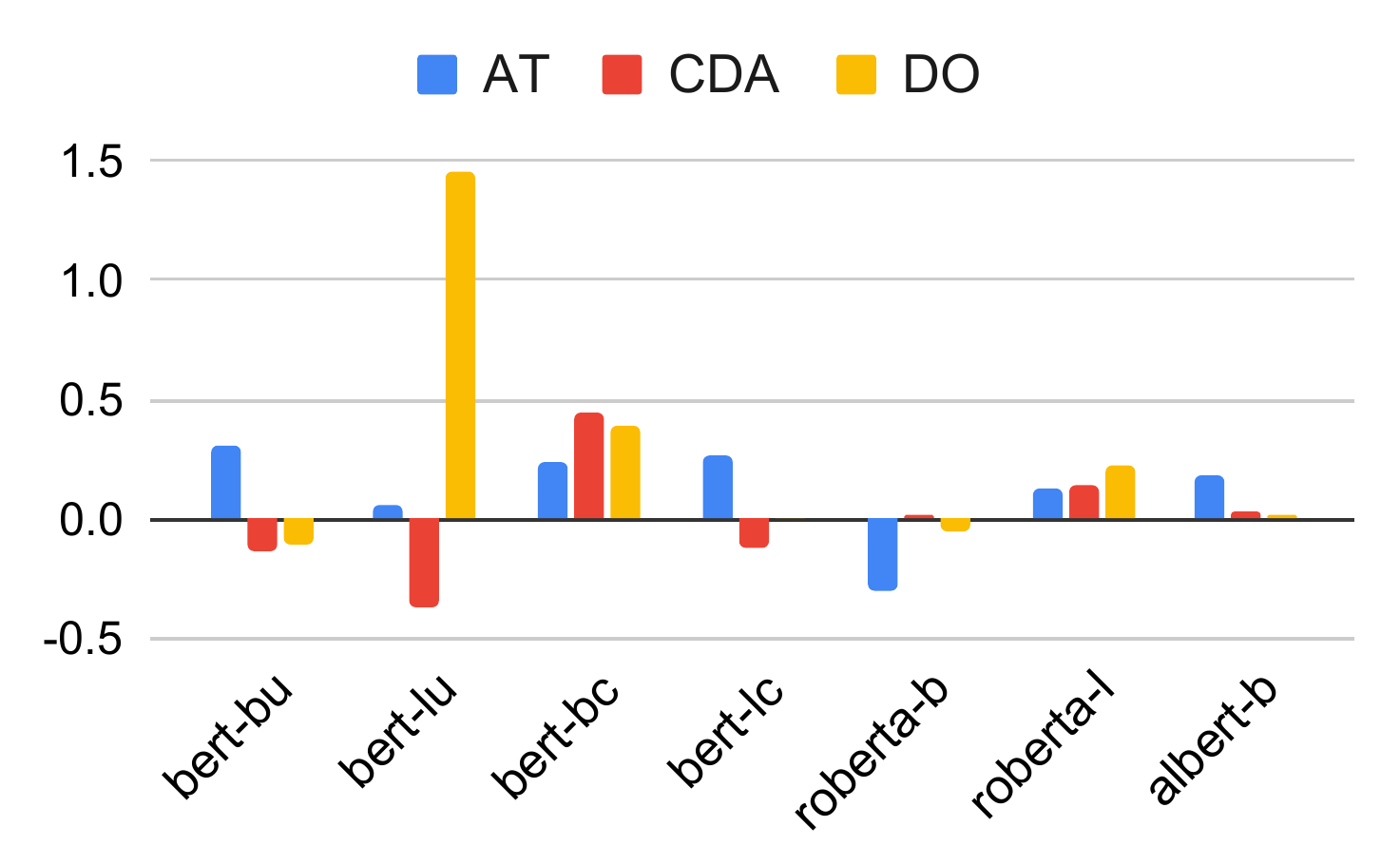}}
    \subfigure[STS-bias]{\includegraphics[clip, width=0.32\textwidth]{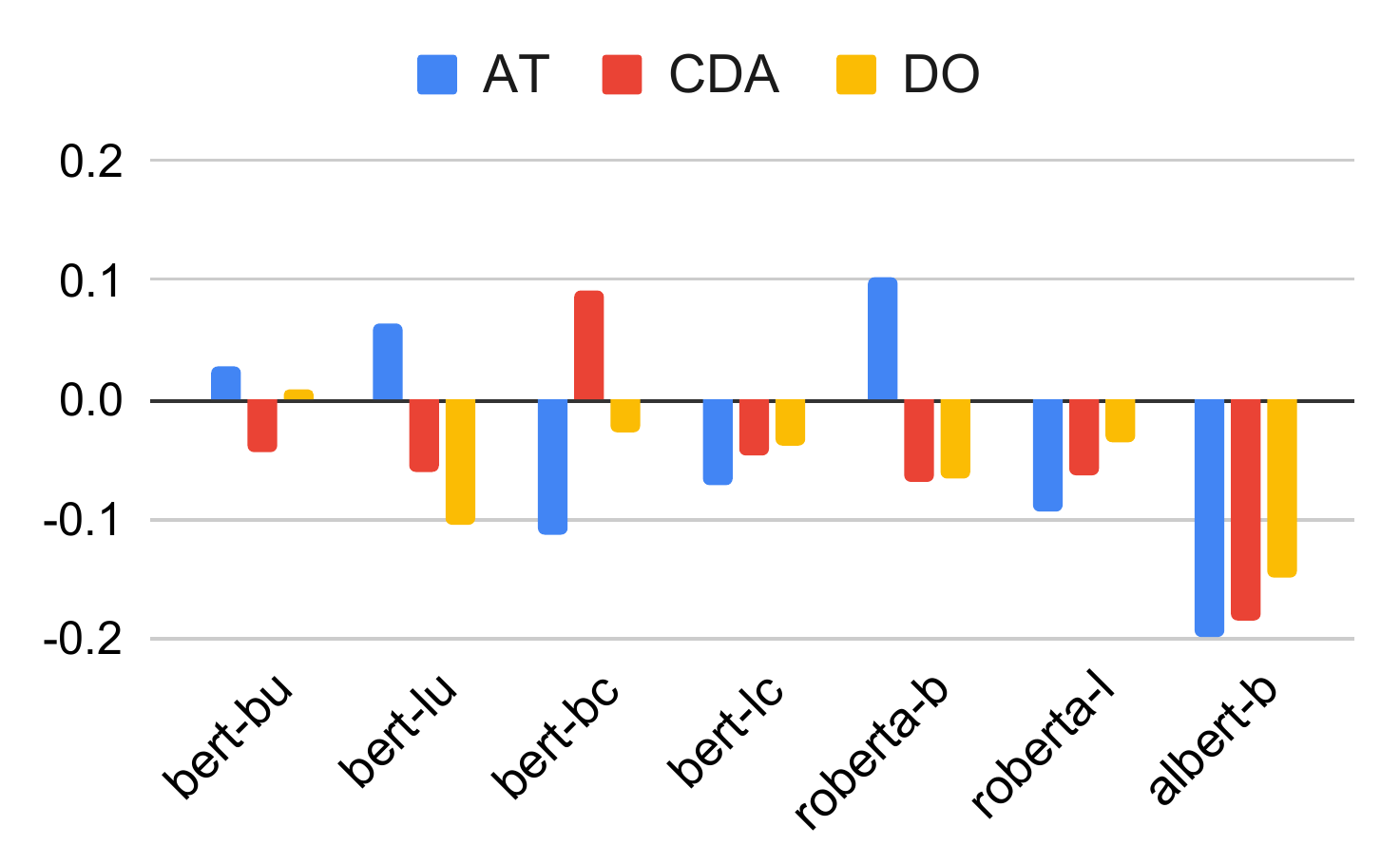}}
    \subfigure[NLI-bias]{\includegraphics[clip, width=0.32\textwidth]{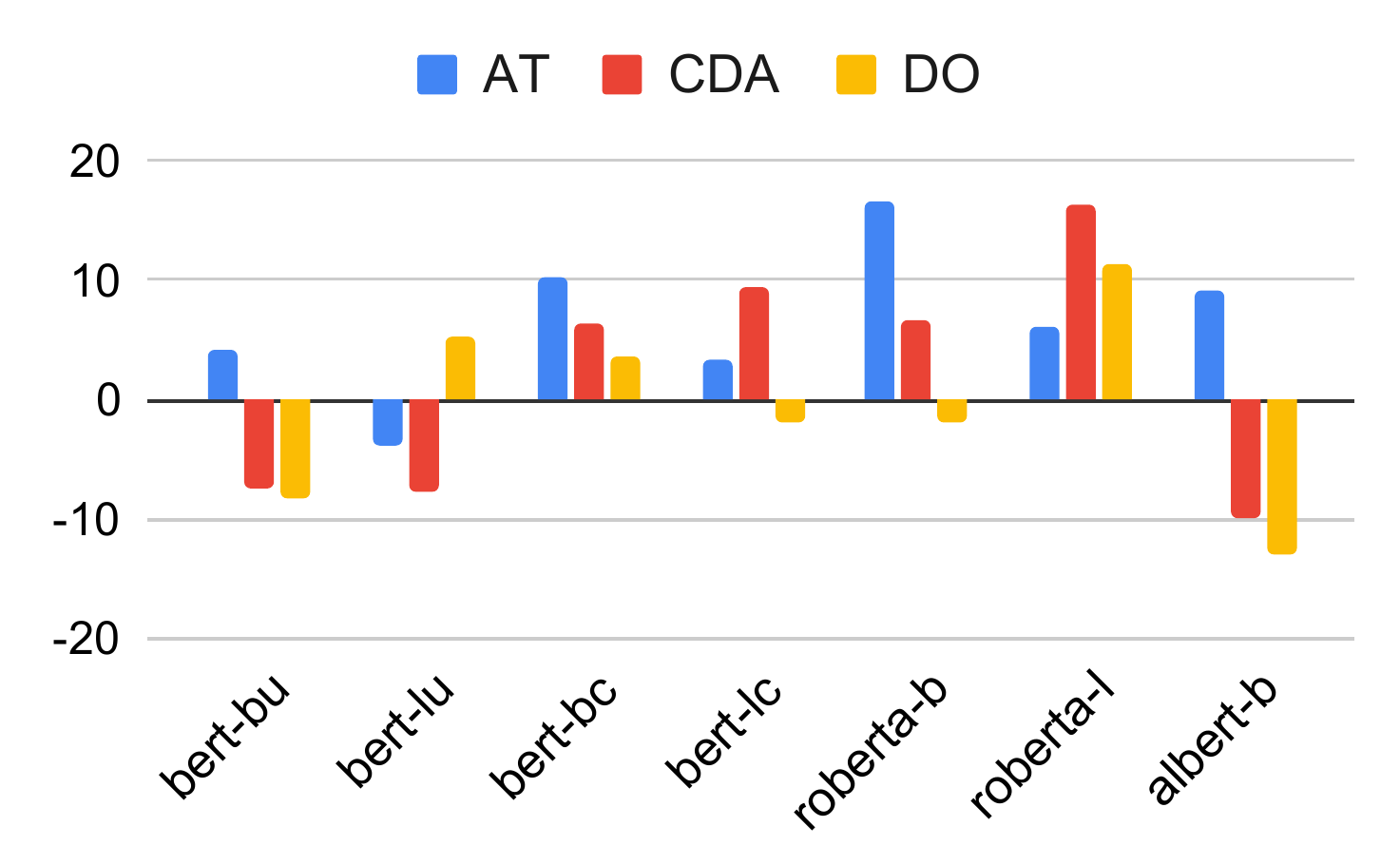}}
\caption{Differences between the bias scores of original vs. debiased MLMs. Negative values indicate that the debiased MLM has a lower bias than its original (non-debiased) version.}
\label{fig:diff}
\end{figure*}

We examine whether the debiasing methods proposed for MLMs can mitigate the social biases when those debiased MLM are used in downstream tasks.
Specifically, given an MLM, $M$, we use each debiasing method described in \autoref{sec:debiasing} independently on $M$ to create a debiased version, $M_{\rm deb}$.
Let, $f(M)$, denote the bias score of $M$ measured using a bias evaluation measure, $f$, described in \autoref{sec:bias-measures}.
\autoref{fig:diff} shows the reduction of bias $f(M_{\rm deb}) - f(M)$ due to debiasing for multiple MLMs and debiasing methods for different $f$.
For all debiasing methods, we see a reduction in biases according to intrinsic bias evaluation measures SSS, CPS and AULA.
However, among the extrinsic bias evaluation measure, we see a reduction according to STS, but an \emph{increase} in biases according to BiasBios and NLI.
For example, AT for \textbf{bert-bu} mitigates bias in all intrinsic measures, but increases biases in all extrinsic measures.
This shows that debiasing methods do \emph{not} always reduces extrinsic biases, despite appearing to be debiased with respect to intrinsic measures.
Note that the training dataset sizes for BioBias, STS-B and MLNI are respectively 251K, 5.7K, and 392.7K.
Therefore, STS-B has fewer training data instances than for the other two tasks, and the number of fine-tuning iterations required is also smaller.
For these reasons, we believe that debiased MLMs are less influenced when fine-tuned on STS-B.

\section{Why is Debiasing less Effective for Downstream Tasks?}

To further investigate the discrepancies between intrinsic and extrinsic bias evaluation measures observed in \autoref{sec:downstream}, 
recall that a debiased MLM can re-learn social biases in two different sources via fine-tuning: \emph{instance-related} biases and \emph{label-related} biases.
A training data point of a downstream task consists of an \emph{instance} (i.e. sentence or text) and a \emph{label} (i.e. entailment labels, similarity ratings etc.).
Instances alone could express social biases such as in the sentence ``\emph{All Muslims are terrorists}''.
By fine-tuning a debiased MLM on such instances using the masking training objective, it can learn racially discriminatory associations.
We call this the instance-related bias in fine-tuning.

Second, even if an instance is unbiased, coupled with its label, the data point could still express a social bias.
For example, the sentence pair (``\textit{A nurse is walking}'', ``\textit{A woman is walking}'') represents an instance, which does not represent a gender bias.
However, if we are told that the first sentence entails the second (i.e. an entailment label assigned to this instance), it will represent a gender bias.
An NLP model can easily learn such label-related biases in an end-to-end training setting by propagating the prediction losses all the way back to the MLM layers.

\begin{table}[!t]
\centering
\small
\begin{tabular}{lrrr}
\toprule
 & Feminine & Masculine & ILSP \\
\midrule
BiasBios    & 20,720 & 35,412 & -0.53 \\
STS-B         & 1,304 & 2,126 & -0.48 \\
MNLI         & 55,588 & 139,575 & -0.91 \\
\bottomrule
\end{tabular}
\caption{Frequency of feminine and masculine words in downstream task data. ILSP represents the degree of bias between feminine and masculine words.}
\label{tbl:gender_freq}
\end{table}



\begin{figure*}[t!]
    \centering
    \subfigure[BiasBios]{\includegraphics[clip, width=0.9\textwidth]{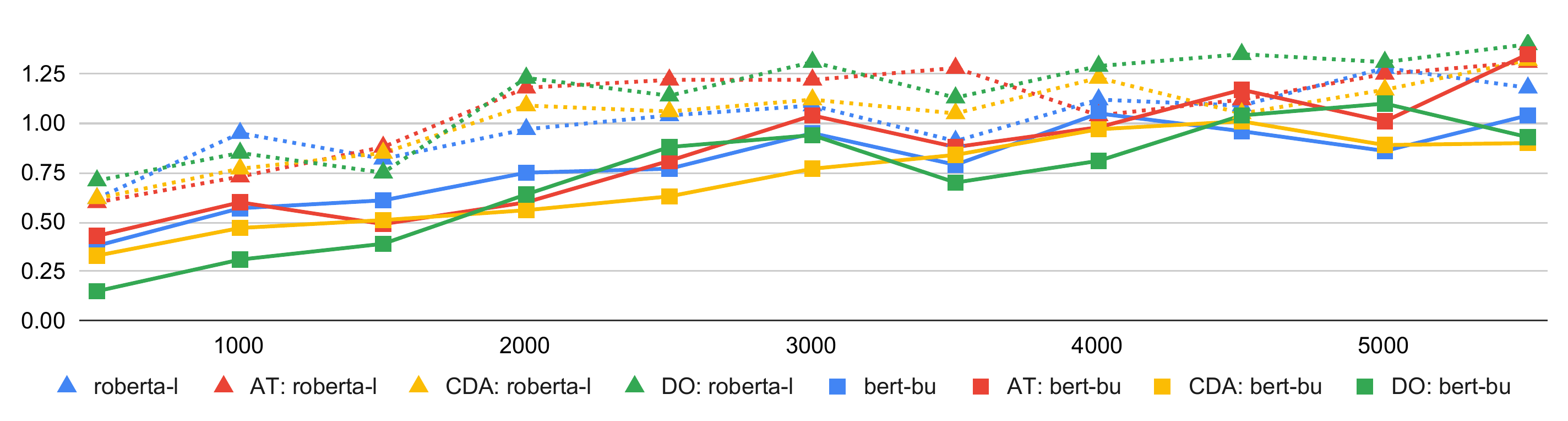}}
    \subfigure[STS-bias]{\includegraphics[clip, width=0.9\textwidth]{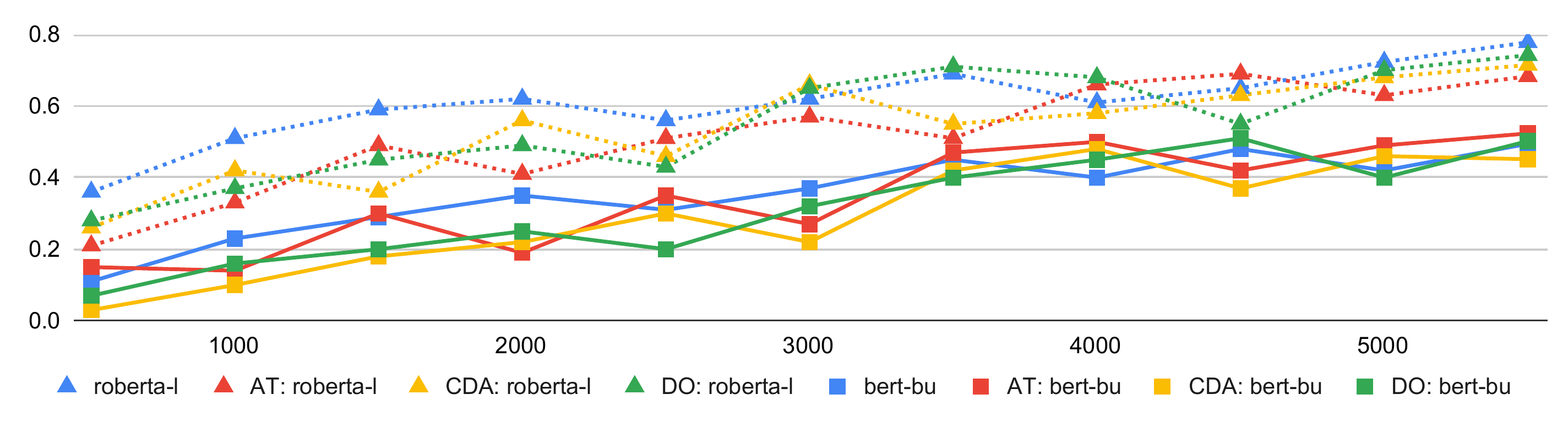}}
    \subfigure[NLI-bias]{\includegraphics[clip, width=0.9\textwidth]{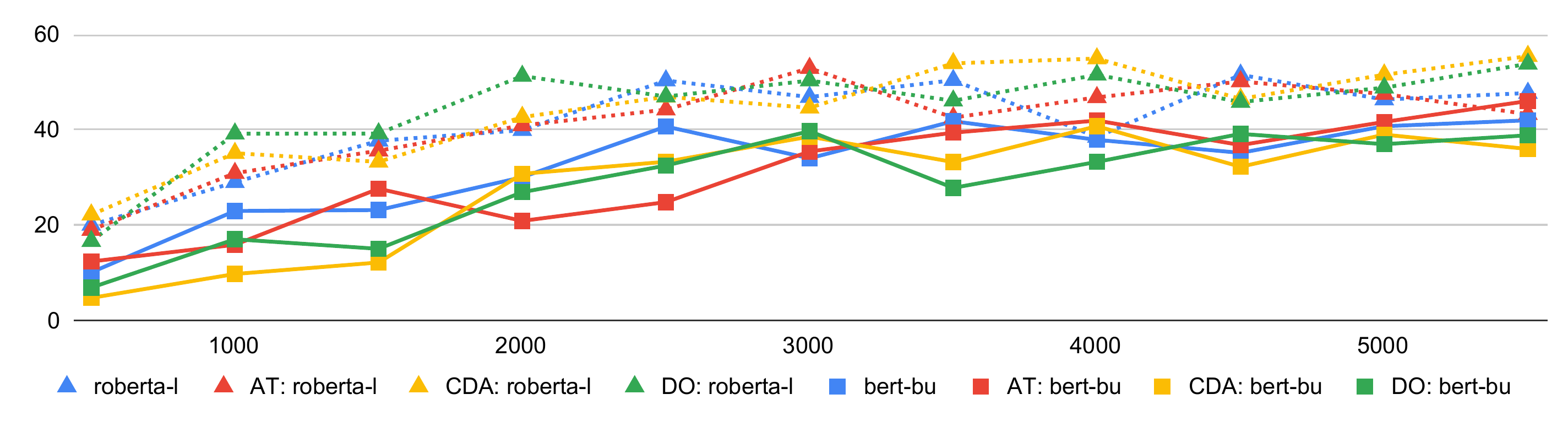}}
\caption{Extrinsic bias scores measured for MLMs during fine-tuning on downstream instances.}
\label{fig:iter}
\end{figure*}

To examine the instance-related gender biases in downstream training data, we compute ILPS~\cite{kurita-etal-2019-measuring} given by $\log(P_f/P_m)$, where $P_f$ and $P_m$ are the probabilities of respectively female and male genders in training instances. 
Prior work has shown that the frequency imbalance of feminine and masculine words in the training data is related to the gender bias learnt by MLMs~\cite{Kaneko:AAAI2022,Kaneko:MLM:2022}.
Using the gender word lists created by \newcite{zhao-etal-2018-learning}, we estimate ILPS as shown in \autoref{tbl:gender_freq}.

From \autoref{tbl:gender_freq}, we see that all three datasets are highly biased towards the male gender, indicated by the negative ILSP scores.
If we arrange the three datasets in the ascending order of their gender biases we get: STS-B < BiasBios < MLNI.
Interestingly, this is exactly the same order we would get if we had ordered these datasets considering the number of MLMs that have their biases increased due to fine-tuning in  \autoref{fig:diff} (e, d, f), respectively.

\subsection{Re-learning Biases via Fine-Tuning}
\label{sec:re-learn}

To further study how biases are learnt during the fine-tuning process, we track the extrinsic bias score of an MLM (and its debiased variants) over the number of fine-tuning training iterations on a particular downstream task.
Here, we use \textbf{bert-bu} and \textbf{roberta-l}, which had low intrinsic biases after debiasing according to all intrinsic bias measures in \autoref{fig:diff}.
We save model checkpoints for every 500 fine-tuning iterations up to 5500 iterations and evaluate their extrinsic bias scores.

\autoref{fig:iter} shows that the extrinsic bias scores of non-debiased (original) MLMs (\textbf{roberta-l} and \textbf{bert-bu}) and their debiased variants using AT, CDA, and DO.
On all downstream tasks, extrinsic bias scores increase during fine-tuning for both non-debiased and debiased MLMs.
This shows that biases that were mitigated by debiasing are re-learnt during fine-tuning on downstream training data.
Even STS-bias, which reported almost all MLMs to be void of biases after debiasing (\autoref{fig:diff}), shows increasing bias scores during fine-tuning.
Overall, we see that debiasing effect gradually decreases as fine-tuning progresses due to the biases in downstream task data.
In contrast, \citet{jin-etal-2021-transferability} argued that debiased MLMs are unaffected by fine-tuning.
However, their evaluation was limited to only one MLM and two datasets both related to gender stereotyping~\cite{jin-etal-2021-transferability}, whereas we consider seven MLMs, three debiasing methods and six bias evaluation measures.
This shows that careful evaluation using various evaluation datasets and MLMs is necessary when verifying the effectiveness of debiasing methods. 

Furthermore, according to the intrinsic evaluations in \autoref{fig:diff}, the biases were reduced in both \textbf{roberta-l} and \textbf{bert-bu} by all debiasing methods.
However, there are cases where the bias scores of debiased MLMs are higher than that of non-debiased MLMs even in the early stages of fine-tuning, except for \textbf{roberta-l} in the STS-bias in \autoref{fig:iter}.
Therefore, MLMs judged to be debiased according to intrinsic evaluations alone might not necessarily remain debiased after fine-tuning on downstream task instances.

\subsection{Correlations between Intrinsic vs. Extrinsic Bias Evaluation Measures}
\label{sec:label-bias}


Given that intrinsic bias evaluation measures are used to determine whether an MLM should be used in a downstream application, it is important to examine what level of correlations exist between intrinsic and extrinsic bias evaluation measures~\cite{blodgett-etal-2021-stereotyping}.
For this purpose, using 28 MLMs (original 7 MLMs and their debiased versions using three methods, $7 \times 3 + 7 = 28$) we compute the Kendall's $\tau$ rank correlation coefficients between intrinsic and extrinsic bias measures.
Specifically, in \autoref{tbl:corr} we show $\tau$ values between the intrinsic bias measure (SSS, CPS, AULA) and extrinsic measures (BiasBios, STS-bias, NLI-bias).
In each cell in \autoref{tbl:corr}, the values to the left of the slashes (/) are computed using the MLMs prior to fine-tuning on downstream data, whereas the values to the right of the slashes are computed using the MLMs that have been fine-tuned on instances of downstream data.\footnote{We use the fine-tuned models after 5500 iterations in \autoref{fig:iter}.}
Statistically significant ($p < 0.05$) $\tau$ values according to the Fisher's exact test are denoted by a $\dagger$.
Cases where the correlation has improved after fine-tuning are are highlighted in \textbf{Bold}.

From \autoref{tbl:corr}, we see that prior to fine-tuning, only two $\tau$ values were significant, reconfirming the weak correlations observed in prior work~\cite{delobelle2021measuring,goldfarb-tarrant-etal-2021-intrinsic,Cao:2022}.
Given that intrinsic bias evaluation measures are agnostic to the biases in downstream task-specific training data, one might have expected to see the correlations between intrinsic and extrinsic measures to improve \emph{after} the MLMs have been fine-tuned on downstream data, thereby capturing some of the biases in downstream data within the MLMs.
Surprisingly, we see that this is clearly not the case here and only in three out of nine combinations in \autoref{tbl:corr} we see correlations improving after fine-tuning.
In particular, for NLI-bias we see that correlations dropping after fine-tuning against all intrinsic measures compared to their values prior to fine-tuning.
On the other hand, for BiasBios we see that correlations improving after fine-tuning against SSS and AULA by large margins, while remaining unaffected against CPS.
The only significant correlation after fine-tuning is between STS-bias and SSS (i.e. 0.38). 
However, this too has decreased from its value prior to fine-tuning (i.e. 0.53).

Overall, our results indicate weak and inconsistent correlations between intrinsic and extrinsic bias evaluation measures irrespective of whether the MLMs have been fine-tuned on downstream data or otherwise.
Therefore, we recommend that intrinsic evaluation scores alone must not be used to determine whether an MLM is sufficiently unbiased to be deployed to NLP systems that are used by billions of users world-wide.

\begin{table}[t!]
\centering
\small
\begin{tabular}{lccc}
\toprule
 & SSS & CPS & AULA \\
\midrule
BiasBios    & -0.02 / \textbf{0.24}    & -0.17 / -0.17 & -0.20 / \textbf{0.11} \\
STS-bias         & 0.53$^\dagger$ / 0.38$^\dagger$     & -0.08 / \textbf{0.12}    & 0.24 / 0.13 \\
NLI-bias         & 0.22 / 0.21 & 0.16 / 0.05     & 0.30$^\dagger$ /0.17\\
\bottomrule
\end{tabular}
\caption{Kendall $\tau$ correlation coefficients between intrinsic extrinsic bias scores. Values to the left and right of '/' are computed using MLMs respectively prior and post fine-tuning. Statistically significant ($p < 0.05$) $\tau$ values are denoted by a $\dagger$, whereas improved correlations after fine-tuning are highlighted in \textbf{Bold}.}
\label{tbl:corr}
\end{table}

    
    

\section{Conclusion}

We investigated social biases in 28 MLMs and found that there exist weak correlation between intrinsic and extrinsic bias evaluation measures.
Moreover, debiased MLMs re-learn biases during task-specific fine-tuning due to instance and label related biases in downstream task data.
Our findings highlight limitations in existing bias evaluation measures and raise serious concerns regarding their implications.


\section{Ethical Considerations}
\label{sec:ethics}

Our goal in this paper was to evaluate the correlation between previously proposed and widely-used intrinsic and extrinsic bias evaluation measures for evaluating various types of social biases in pretrained MLMs.
However, we did not manually annotate novel social bias datasets or proposed novel bias evaluation measures nor debiasing methods.
Therefore, we do not see any ethical issues arising due to data annotation, or via proposals of novel evaluation metrics or debiasing methods.
However, we would like to point out that although we conducted an extensive study using 7 MLMs, 3 debiasing methods, 3 intrinsic evaluation measures and 3 extrinsic evaluation measures, this is still a limited subset of all possible scenarios.
For example, the gender biases we considered in this paper cover only binary gender~\cite{dev-etal-2021-harms}.
However, non-binary genders are severely underrepresented in textual data used to train MLMs.
Moreover, non-binary genders are often associated with derogatory adjectives.
Evaluating the correlation between intrinsic and extrinsic non-binary gender is an important next step.

All MLMs used in this study are trained for English.
However, as reported in much prior work social biases are language independent and omnipresent in MLMs trained for many languages~\cite{Kaneko:MLM:2022,lewis2020gender,liang-etal-2020-monolingual,bartl-etal-2020-unmasking,zhao-etal-2020-gender}.
We plan to extend this study to cover non-English MLMs in the future.

\citet{kaneko2022gender} show that combining multiple debiasing methods for word embeddings~\cite{Tolga:NIPS:2016,kaneko-bollegala-2019-gender,kaneko-bollegala-2021-dictionary,kaneko-bollegala-2020-autoencoding,ravfogel-etal-2020-null} can complement each other’s strengths and weaknesses to obtain more effective debiased embeddings.
Therefore, combining methods to mitigate both intrinsic and extrinsic bias may allow for more effective debiasing.

Finally, biases are not limited to word representations but also appear in sense representations~\cite{Zhou:2022}.
On the other hand, our correlation analysis did not include any sense embedding models.

\bibliography{custom}
\bibliographystyle{acl_natbib}

\end{document}